\documentclass[lettersize,journal]{IEEEtran}
\usepackage{amsmath,amsfonts}
\usepackage{algorithmic}
\usepackage{algorithm}
\usepackage{array}
\usepackage[caption=false,font=normalsize,labelfont=sf,textfont=sf]{subfig}
\usepackage{textcomp}
\usepackage{stfloats}
\usepackage{url}
\usepackage{verbatim}
\usepackage{graphicx}
\usepackage{cite}
\usepackage{float}
\usepackage{stfloats}
\usepackage{url}
\usepackage{booktabs}       
\usepackage{amsfonts}       
\usepackage{nicefrac}       
\usepackage{microtype}      
\usepackage{xcolor}         
\usepackage[table,xcdraw,dvipsnames]{xcolor}
\usepackage{multirow}
\usepackage{amsmath}
\usepackage{xspace}
\usepackage{graphicx}       
\usepackage{wrapfig}
\usepackage{tabularx}
\usepackage{makecell}
\usepackage{adjustbox}
\usepackage{array}
\usepackage{epsfig}
\usepackage{times}
\usepackage{amssymb}
\usepackage{caption}
\usepackage{textcase}
\usepackage{threeparttable}
\usepackage{pifont}

\colorlet{colorFst}{Green!25}       %
\colorlet{colorSnd}{SpringGreen!50} %
\colorlet{colorTrd}{Yellow!35}      %

\makeatletter
\DeclareRobustCommand\onedot{\futurelet\@let@token\@onedot}
\def\@onedot{\ifx\@let@token.\else.\null\fi\xspace}

\def\eg{\emph{e.g}\onedot} 
\def\ie{\emph{i.e}\onedot} 
 
\def\etc{\emph{etc}\onedot}

\makeatother
\begin{document}

\title{Compact Object-Level Representations with Open-Vocabulary Understanding for Indoor Visual Relocalization}

\author{ Zhaopeng Cui$^{1}$, Jiarui Hu$^{1\dag}$, Jingbo Liu$^{1}$, Boming Zhao$^{1}$, Xiyue Guo$^{1}$, Boyin Feng$^{1}$, Haocheng Peng$^{1}$, \\Yujun Shen$^{2}$, Hujun Bao$^{1}$, Guofeng Zhang$^{1\dag}$

\thanks{Manuscript received February 6, 2026; Revised May 4, 2026; Accepted June 3, 2026.}
\thanks{This paper was recommended for publication by Editor Ayoung Kim upon evaluation of the Associate Editor and Reviewers’ comments. This work was partially supported by the National Key R\&D Program of China (Grant No. 2024YFB4505500 \& 2024YFB4505501), the NSFC (No.~62572425 and No.~624B2132), and Ant Group.}
\thanks{\dag~Corresponding Authors}
\thanks{$^{1}$State Key Lab of CAD\&CG, Zhejiang University, $^{2}$Ant Group}
}

\markboth{IEEE ROBOTICS AND AUTOMATION LETTERS. PREPRINT VERSION. ACCEPTED JUNE 2026}%
{Shell \MakeLowercase{\textit{et al.}}: A Sample Article Using IEEEtran.cls for IEEE Journals}


\maketitle

\begin{abstract}
Indoor visual relocalization plays a critical role in emerging spatial and embodied AI applications. However, prior research was predominantly devoted to low-level vision schemes, struggling to perceive scene semantics and compositions, which limits both interpretability and applicability. In this paper, we explore the issue of how to organize rich object information in a scene, including semantics, layout, and geometry, into a structured map representation, thereby utilizing object units exclusively to drive the camera relocalization task. To this end, we propose OpenReLoc, a camera relocalization system designed to provide scene understanding and accurate pose estimation capabilities. Leveraging recent foundation models, we first introduce a multi-modal mechanism to integrate open-vocabulary semantic knowledge for effective 2D-3D object matching. 
Additionally, we design object-oriented reference frames as position priors, paired with a reference frame selection strategy based on the Distance-IoU (DIOU), enabling extension to scalable scenes.
Moreover, to ensure stable and accurate pose optimization, we also propose a dual-path 2D Iterative Closest Pixel loss guided by object shape. Experimental results demonstrate that OpenReLoc achieves superior relocalization recall and accuracy across various datasets. Our source code will be released upon acceptance.
\end{abstract}

\begin{IEEEkeywords}
Open-vocabulary Scene Understanding, Camera Relocalization, Object-level Representation
\end{IEEEkeywords}

\section{Introduction}
\label{sec:intro}
\IEEEPARstart{I}{ndoor} camera relocalization has been a fundamental problem in 3D computer vision for decades, especially with trending applications such as virtual/augmented reality (VR/AR), robot-environment interaction, and navigation. The goal is to estimate the 6-DOF camera pose given a visual observation in a known map. At present, facing increasingly challenging requirements for embodied agents, an indoor relocalization system is desired to evolve beyond just accuracy, towards scalability, compactness, and most importantly, semantic awareness, to improve its versatility and adaptability in various situations.

Previous visual relocalization methods~\cite{coordinet, pixloc, mur2017orb, ms-transformer} mainly rely on low‐level visual features, and thus suffer from limitations in robustness, compactness, and semantic awareness. Overlooked by these methods, an indoor scene is essentially a spatial arrangement of 3D objects, characterized by rich semantics, regular geometry, and a distinct layout. Low-level appearance cues will degrade under challenging conditions, like illumination changes, whereas object entities and their spatial relations can still remain stable. Low-level visual features typically need to be attached to dense points in a point cloud, resulting in heavy map overhead, while object-level representations associate one embedding per object. This naturally yields a more compact map~\cite{zins2022oa, goreloc}. In addition, downstream tasks reason over semantic objects rather than raw feature points. An object-level map explicitly captures objects with rich semantic properties, naturally aligning with how robots plan and act~\cite{peng2023openscene}. Poses relocated from such a map convey interpretable spatial relationships between a robot and its surroundings (e.g., `next to the sofa and facing the table'), providing valuable context for navigation. Given these advantages, using an object-level map for camera relocalization is a promising solution that achieves a holistic balance across multiple performance aspects, especially for indoor scenarios rich in numerous objects.

However, there are only a few emerging attempts~\cite{goreloc} to exploit objects in camera relocalization, and their system designs remain underdeveloped. Specifically, they are limited by three primary drawbacks: \textbf{(1)} Existing landmark association techniques employ object descriptors with low discriminability, which leads to severe outliers in object matching. \textbf{(2)} Reliable pose priors are necessary for relocalization in scalable indoor scenes, yet they are absent in existing object-level works. \textbf{(3)} Previous works optimize camera poses via aligning 2D-3D bounding box centers, which is significantly ambiguous and error-prone under sparse object correspondences. A dedicated pose optimization strategy tailored to the object-level paradigm is still lacking.

In response to these challenges, we propose \textbf{OpenReLoc}, a semantic-aware, memory-efficient, and scalable camera relocalization framework based on object-level representations with open-vocabulary understanding. 
We construct an object-oriented map suite that consists of a global scene graph, open-vocabulary object descriptors, object geometry, and reference frames.
First, to overcome the landmark association issue, our system leverages open-vocabulary descriptors and the global scene graph to perform multi-modal object matching.  We utilize an off-the-shelf foundation model, CLIP~\cite{clip}, to embed both visual and textual concepts into object descriptors, which capture high-level semantic knowledge such as affordance, material, \etc, enabling recognition of class-agnostic objects. Additionally, the global scene graph provides layout context as an informative modality to further enhance landmark association.
Second, to extend to scalable scenes, we design object-oriented reference frames, a concise format that records only observed object IDs and 2D bounding boxes instead of full RGB images. Building on this design, a DIOU-based (Distance-IOU) retrieval strategy is also derived to measure frame similarity between query and database images, providing reliable pose priors.
Third, to improve object-level pose optimization accuracy, we propose a dual-path 2D ICP (Iterative Closest Pixel) loss to align observed and actually projected pixel areas of objects. This fine-grained alignment can provide stable pose guidance even under sparse object correspondences. Benefiting from this loss, OpenReLoc achieves exceptional accuracy gain over previous solutions.

\begin{figure*}[t]
    \centering
    \includegraphics[width = 0.93\linewidth]{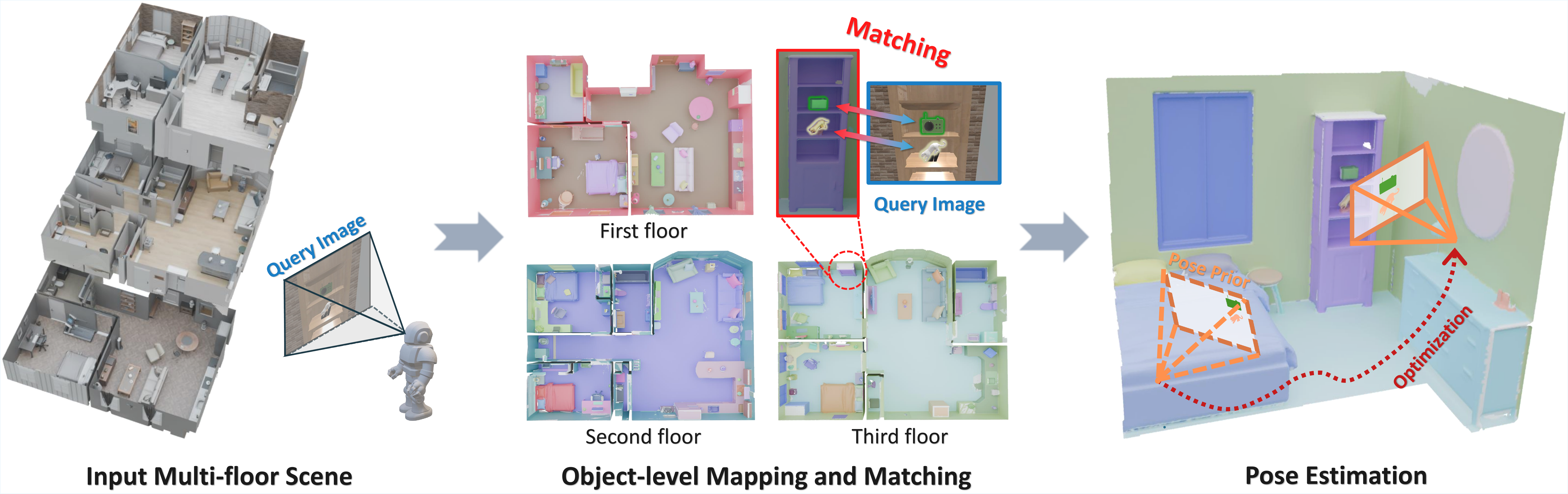}
    \caption{\textbf{OpenReLoc}, an open-vocabulary visual relocalization system, can achieve robust and accurate relocalization performance on various indoor scenes, based on an object-level map. As shown in the figure, in an extremely large multi-floor scene, the robot observes a tiny corner containing a \textcolor[rgb]{0,0.7,0}{small radio} and \textcolor[RGB]{159,145,61}{long-tailed animal ornament}, and our system successfully identifies their 3D correspondences from hundreds of landmarks in the map. Next, initializing a pose prior, we can refine the camera pose with a novel loss.}
    \vspace{-1.5em}
\label{fig:teaser}
\end{figure*}

We evaluate our system on indoor benchmarks, including ScanNet~\cite{scannet} and ScanNet++~\cite{scannet++}. More importantly, we synthesized multiple large-scale scenes based on the Habitat~\cite{habitat19iccv} simulator to cover a wide range of object categories. To the best of our knowledge, OpenReLoc is the first object-level system that can work in such large-scale scenes. (See in Fig.~\ref{fig:teaser}). Extensive experiment results demonstrate that our system outperforms existing approaches, yielding superior recall and accuracy in visual relocalization. Overall, our contributions can be summarized as follows:

\begin{itemize}
    \item We introduce a multi-modal landmark association module that combines open-vocabulary object descriptors with a global scene graph, enabling robust class-agnostic object matching.
    \item We design object-oriented reference frames and derive a DIOU-based retrieval strategy together to provide reliable pose priors in scalable scenes.
    \item We propose a dual-path 2D ICP loss for pose optimization, delivering stable pose guidance and improved accuracy even with sparse object correspondences.
    \item OpenReLoc presents a comprehensive framework with an object-level map suite, remaining semantics, robustness, accuracy, and compactness. Experiments on various datasets demonstrate the excellent system performance.
\end{itemize}

\section{Related Work}
\label{sec:relatedwork} 
\noindent\textbf{Open-Vocabulary Semantic Mapping.}  Traditional semantic mapping commonly trains a neural classifier on fixed object categories. While effective on known scenes and objects, they fail to generalize to long-tail categories and complex scenarios. Recent progress in 2D vision-language foundation models, such as~\cite{clip,jia2021scaling}, has advanced the shift in semantic mapping from closed-set approaches to open-vocabulary ones. This enables robust zero-shot recognition and alleviates the need for labor-intensive annotations. OpenScene~\cite{peng2023openscene} introduces the open-vocabulary scene mapping and understanding task. It directly projects 2D CLIP features onto dense 3D point clouds. However, point-wise feature alignment is prone to noisy and incomplete segmentation results. \cite{takmaz2023openmask3d,nguyen2024open3dis} adopt a 3D model to first generate class-agnostic instance proposals, then extract instance-wise open-vocabulary features, which improves the accuracy of object recognition. Nevertheless, pre-trained 3D models struggle to ensure reliable segmentation performance. Following \cite{yan2024maskclustering,lu2023ovir} utilize powerful 2D segmentation models to produce 2D class-agnostic masks, and merge them into instances. Lifting 2D segmentations into 3D space can effectively enhance instance quality. The above works provide references on how to build an object-oriented map for object-level relocalization.

\noindent\textbf{Object-Level Relocalization.}
Recently, object-level SLAM has gained widespread attention. By matching the mapping frames with 3D instances and minimizing the projection error, object-level SLAM~\cite{zins2022oa,yang2019cubeslam} shows satisfactory pose results. Although object-based SLAM methods have been researched, object-based relocalization remains relatively underexplored. Object matching is prone to incorrect associations, which may lead to degradation in relocalization accuracy. GOReloc~\cite{goreloc} considers the semantic uncertainty and consistency in a graph to facilitate object matching. But such graph-based object descriptors record only closed-vocabulary neighbor categories and counts, and thus remain weak in the landmark association. Clip-Loc~\cite{clip-loc} first tried to introduce open-vocabulary features as object descriptors, and Clip-Clique~\cite{CLIP-Clique} further proposed to combine maximal clique finding to improve matching performance. Unfortunately, they lacked a complete map suite and a systematic pipeline, which prevented them from fully exploring the potential of this research line. This is also the reason for their limited performance and scalability.

\section{Method}
\label{sec:method}

\begin{figure*}[t]
    \centering
    \includegraphics[width = 0.8\linewidth]{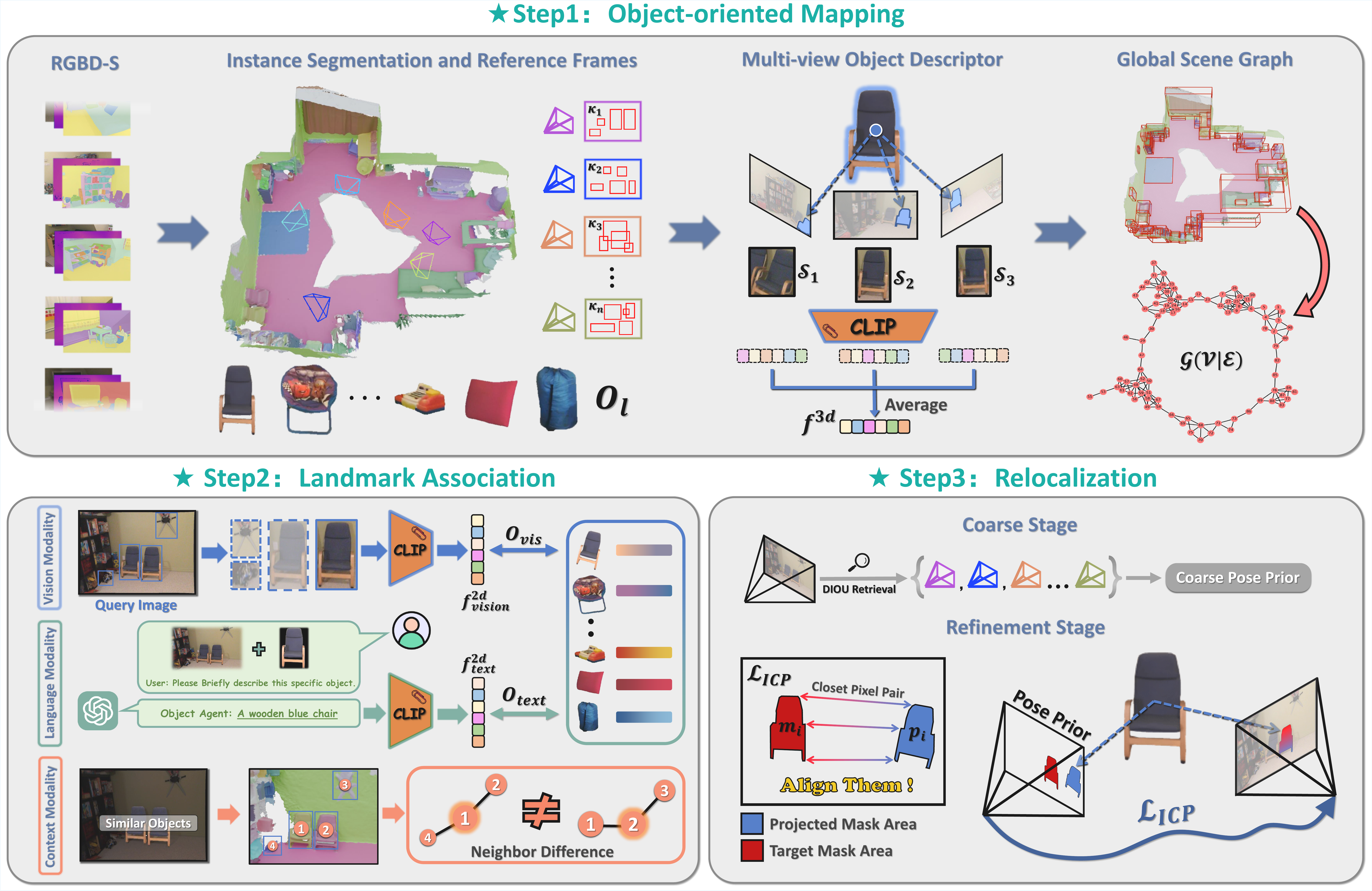}
    \caption{\textbf{System Overview.} Our system includes three main steps: (1) Object-oriented Mapping. We construct an object-level map from an RGB-D sequence and its 2D segmentations, comprising object landmarks $O_l$, descriptors $f^{3d}$, reference frames $\mathcal{K}$, and a global scene graph $\mathcal{G}$. (2) Landmark Association. Given a query image, multi-modal features ${f^{2d}_{vision},f^{2d}_{text}}$ are extracted and matched against $f^{3d}$, accompanied by scene graph analysis. (3) Relocalization. A coarse pose prior from DIOU-based retrieval can be further refined using the 2D ICP loss.}
    \label{fig:pipeline}
    \vspace{-1.5em}
\end{figure*}

\noindent \textbf{Problem Statement.} The overview of our proposed object-level camera relocalization framework is shown in Fig~\ref{fig:pipeline}. Given a collection of posed RGBD images from a scene, and an unseen query RGB image from the same scene, the topic of the object-level camera relocalization task is to estimate the 6-DoF camera pose of this query image solely based on key attributes of object-level representations, such as semantics, neighbor relationships, and geometric shapes.

\noindent \textbf{Step-by-step Overview.} \textit{Object-oriented Mapping} (Sec~\ref{sec3-1}): Given a set of posed RGBD images from a scene, this step is to process these RGBD observations and build an object-centric 3D map suite, including 3D instance segmented point clouds, per-object feature descriptor, object-oriented reference frames, and a global scene graph. \textit{Landmark Association} (Sec~\ref{sec3-2}): Given the pre-built map from the last step and an unseen RGB query image, we analyze this query image and establish correspondences between observed objects in the query image and those in the 3D map. \textit{Relocalization} (Sec~\ref{sec3-3}): Given the 3D map, 2D query image, and object matching pairs between them, we employ a coarse-to-fine strategy to accurately and robustly estimate the 6-DoF camera pose of this query image.

\subsection{Object-oriented Mapping}
\label{sec3-1}
Object-oriented mapping is the first and pivotal step in our framework, where a well-structured map suite and high-quality reconstruction form the core foundation. In this section, we introduce an object-oriented mapping workflow and the principles behind each module.

\noindent \textbf{Instance Segmentation.} In the object-level relocalization task, instance segmentation plays an important role in identifying individual objects. Based on depth observations, we can reconstruct the scene mesh by TSDF-Fusion~\cite{zeng20173dmatch} and convert vertices into the scene point cloud $\mathcal{P}$. 
Then, as in~\cite{yan2024maskclustering}, we predict 2D mask proposals on RGB images as nodes and compute their multi-view consensus as edge affinity. Through graph clustering, we can merge nodes into clusters, each representing an instance.
Instance segmentation module lifts 2D segmentations into 3D space and generates landmarks $\{O_l^i=(P_i,B^{3d}_i,C_i)|i=1,2,3..N\}$ including point clouds $P_i$, 3D bounding boxes $B^{3d}_i$, and centers $C_i$.

\noindent \textbf{Distribution of Reference Frames.} In order to adapt to scalable scenes, we design object-oriented reference frames and distribute them as initial pose anchors. Unlike the RGB reference frame, our design only records observed landmark IDs $i$ and 2D bounding box coordinates $B^{2d}_i$ instead of full RGB images. We show our reference frames $\mathcal{K}$ in Fig.~\ref{fig:pipeline} and define them as Eq.~\eqref{eq-2}.
\begin{equation}
    \mathcal{K} = \{(i,B^{2d}_i)|i=1,2,..,N_\mathcal{K}\}~,
    \label{eq-2}
\end{equation}
where $N_\mathcal{K}$ represents the number of objects in $\mathcal{K}$. The visibility of an object is measured by projecting its point cloud onto the image plane and calculating the ratio of points within the frame boundaries. Reference frames are added under two conditions: when a new object appears, or when an object's visibility exceeds twice the maximum observed in previous reference frames. This empirically determined criterion provides a favorable trade-off between the number of reference frames and the quality of coarse pose priors. Such object-oriented reference frames can help mitigate storage overhead caused by RGB images, especially in large-scale indoor scenarios.

\noindent \textbf{Multi-view Object Descriptor.} Recent progress suggested that the advanced CLIP model can work as an effective object descriptor encoder~\cite{peng2023openscene}. We project point clouds $P_i$ on the image plane to find patches $\mathcal{S}$ of the same landmark in different views. Top-$k$ patches with maximal visibility are input into a CLIP visual encoder and an average pooling layer to obtain a multi-view CLIP feature $f^{3d}$:
\begin{equation}
    f^{3d}_i = \frac{1}{k} \sum_{n=1}^{k} \mathrm{CLIP}(\mathcal{S}_n)~.
    \label{eq-1}
    \vspace{-0.2em}
\end{equation}
Rich open-vocabulary semantics encoded in $f^{3d}$ enable it to act as discriminative descriptors for class-agnostic objects.

\noindent \textbf{Invalid Object Filtering} It is observed that some indoor objects, such as walls and floors, provide little valuable information but occupy a large portion of the map. They are considered invalid or even negative and should be discarded in subsequent steps. For computational simplicity, we identify them by their occurring frequency in reference frames, since objects that appear in most reference frames cannot assist in narrowing the pose retrieval region.

\noindent \textbf{Scene Graph Extraction.} Spatial arrangement is also an essential cue for indoor object-level relocalization. As a typological representation, a scene graph can clearly describe object relationships and the contextual layout to disambiguate the landmark association. In our global scene graph $\mathcal{G}(\mathcal{V}|\mathcal{E})$, edges $\mathcal{E}$ are added from each object node $\mathcal{V}$ to its nearest neighbors or any objects whose 3D bounding boxes overlap.

\subsection{Landmark Association}
\label{sec3-2}
Landmark association refers to associating observed 2D objects in the query image with their counterparts in a pre-built object-level 3D map. This has historically been challenging, especially in large-scale scenes, because previous closed-vocabulary methods exhibit poor generalization in object matching, and their object descriptors are unable to encode multi-modal information outside of category labels.

\noindent \textbf{Open-vocabulary Matching.}
For a query image, we detect object regions within it, and these found regions (2D bounding boxes) are fed into CLIP to obtain visual features $f^{2d}_{vision}$. We can quantify the uniqueness $\gamma$ of a region with its cosine similarity variance with all landmarks, as Eq.~\eqref{eq-3}.
\begin{equation}
    \gamma = Var(cos(f^{2d}_{vision}, f^{3d}_{1}), ...,cos(f^{2d}_{vision}, f^{3d}_{N}))~,
    \label{eq-3}
\end{equation}
where $Var(\cdot)$ is the variance and $cos(\cdot)$ is the cosine similarity. Only unique 2D objects will be retained for subsequent matching.

Next, we hope to automatically produce text descriptions on these unique 2D objects $O_q$. Large Language Models (LLMs) have exhibited remarkable common-sense reasoning ability, motivating us to use an LLM-based agent for object annotations. Nevertheless, we still need a well-crafted LLM prompt to raise response quality. It is well known that an object's functionality is closely tied to its surrounding environment. However, excessive surrounding information may lead to visual interference, negatively impacting LLM inference. Consequently, as in Fig.~\ref{fig:pipeline}, we pass two images into the agent: the query image and a segmented 2D object. This setup encourages the agent to perform more precise analysis. LLM responses are similarly fed into CLIP for language descriptors $f^{2d}_{text}$. We can now correlate multi-modal features ${f^{2d}_{vision},f^{2d}_{text}}$ with $f^{3d}$ for Top-3 landmark candidates $O_{vis}, O_{text}$ respectively:

\begin{subequations}
    \vspace{-1em}
    \begin{equation}
        cos(f^{2d}_{vision}, f^{3d}_{i=1,..,N}) \Rightarrow   O_{vis}\text{=}\{O_l^{v1}, O_l^{v2},O_l^{v3}\}~,
        \label{eq-4a}
    \end{equation}
    \begin{equation}
        cos(f^{2d}_{text}, f^{3d}_{i=1,..,N})  \Rightarrow  O_{text}\text{=}\{O_l^{t1}, O_l^{t2},O_l^{t3}\}~.
        \label{eq-4b}
    \end{equation}
\end{subequations}

If visual and language cues indicate the same 3D object ($O_l^{v1}=O_l^{t1}$), it is regarded as confident enough to be inserted into the final landmark association results $L$. Otherwise, we give a set of 3D object candidates $U\text{=}\{(O_{vis}\cap O_{text})\cup O_l^{v1} \cup O_l^{t1}\}$ to be further checked in scene graph analysis.

\begin{figure}[t]
    \centering
    \includegraphics[width = 0.85\linewidth]{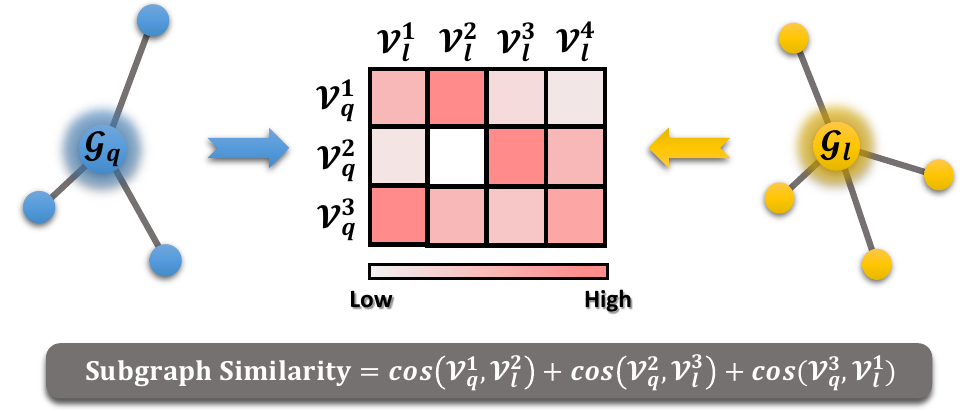}
    \caption{\textbf{Subgraph Similarity.} We seek an assignment among all possible neighbor pairs to maximize the total matching score as the subgraph similarity.}
    \label{fig:subgraph similarity}
    \vspace{-1.5em}
\end{figure}

\noindent \textbf{Sub-graph Matching.} As in Fig.~\ref{fig:pipeline}, frequent repeated or similar objects in an indoor scene are the root cause of the uncertainty set $U$. We resort to neighborhood relationships in the global scene graph to address this problem. For a 2D object in the query image, its subgraph $\mathcal{G}_q$ will be constructed by connecting the nearest and other intersecting 2D object regions. For a candidate in $U$, it is regarded as an origin in the global scene graph $\mathcal{G}(\mathcal{V}|\mathcal{E})$ to extract a Breadth-First 3D subgraph $\mathcal{G}_l$ with a path length $\eta$. We can determine the final candidate from $U$ by finding a 3D subgraph $\mathcal{G}_l$ that is most similar to the $\mathcal{G}_q$.

Another problem is how to measure subgraph similarity, which is formulated as a linear sum assignment problem (LSAP). The goal is to solve an optimal assignment from $\mathcal{G}_q$ to $\mathcal{G}_l$ so that the total matching score is maximized, as shown in Fig.~\ref{fig:subgraph similarity}. After filtering via subgraph matching, we can obtain the final landmark association results:
\begin{equation}
    L = \{(O_q^{i_1}, O_l^{i_2})|i_1\in(1,..,N_q), i_2\in(1,...N)\}~.
    \label{eq-5}
    \vspace{-0.5em}
\end{equation}

\subsection{Relocalization}
\label{sec3-3}
As the final pose estimation step, the object-level tracker directly influences the relocalization success rate and accuracy. Our object-level tracker improves relocalization performance relying on a coarse-to-fine strategy and a novel loss. Two relocalization stages are detailed below.

\begin{figure}[t]
    \centering
    \includegraphics[width = \linewidth]{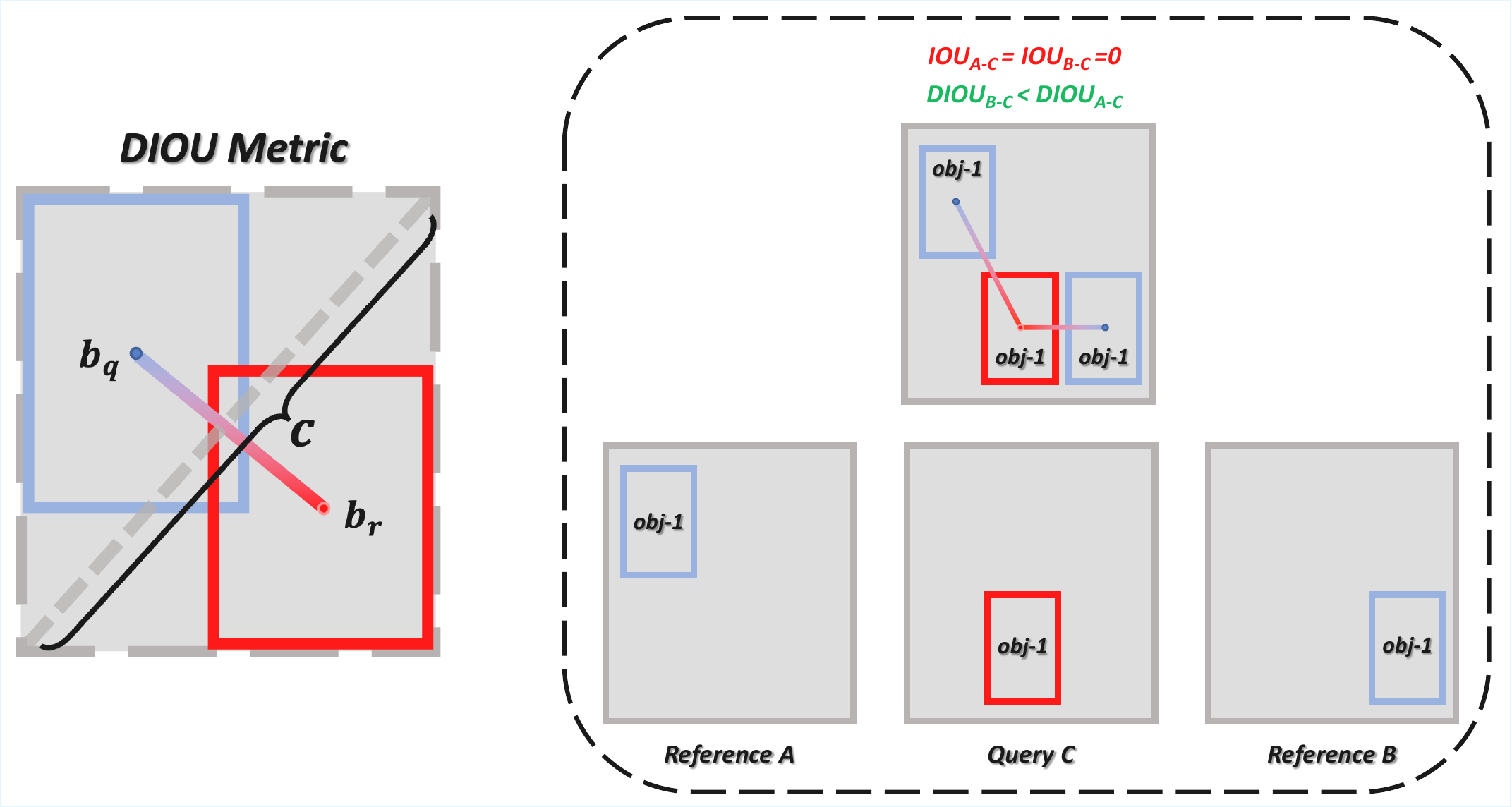}
    \caption{\textbf{DIOU Metric.} (Left) DIOU calculation. (Right) A case illustrates the intention behind the DIOU metric.}
    \label{fig:diou}
    \vspace{-1.5em}
\end{figure}

\noindent \textbf{Coarse Pose Prior.} To vote for a reference frame most similar to the query image, we initially choose the ones that contain the largest number of matched landmarks. This often results in a co-visible subset of reference frames, for which a DIOU metric is further calculated as follows:
\begin{equation}
    \mathrm{DIOU} = 1 - IOU + \frac{||\mathbf{b_q} - \mathbf{b_r}||^2}{c^2}~,
    \label{eq-6}
\end{equation}
where $\mathbf{b_q}$ and $\mathbf{b_r}$ represent 2D bounding box centers of the same object in the query and reference frames, respectively, and $c$ is the diagonal distance of the smallest enclosing rectangle covering two boxes, as shown in Fig.~\ref{fig:diou} (Left). This metric is compatible with our object-level map, and it can avoid some corner cases from non-overlapping boxes, as shown in Fig.~\ref{fig:diou} (Right). The lower the DIOU metric, the better. So far, the reference frame $\mathcal{K}$ with the best DIOU score offers a coarse pose prior.

\noindent \textbf{Refined Pose Optimization.} Previous object-level works always optimize poses by aligning the center points of 2D and 3D bounding boxes. Although this approach can provide a rough estimation, it is inherently unsuitable for an object-level system. In principle, it is prone to severe errors or ambiguity due to extremely sparse correspondences of object center points. Inspired by the traditional 3D ICP (Iterative Closest Point) algorithm, we evolved it into a 2D ICP variant on the image plane to optimize camera poses $\{q, T\}$, \ie, quaternion rotation and translation. Specifically, for the point cloud $P_i$ of a landmark and its target mask area $m_i$ in the query, we project $P_i$ into a pixel set $p_i$ at the current pose and compute the bidirectional average distance between the closest pixel pairs in $p_i$ and $m_i$, as shown in Eqs.~\ref{eq-7a} and \ref{eq-7b}. To ensure more robust and accurate pose estimation, we use a Huber kernel $\mathcal{H}$ with a threshold $\delta$ on the 2D ICP loss to suppress outlier pixels. 
\begin{subequations}
\begin{equation}
    \mathcal{L}^i_{forward} = \frac{1}{N_{p_i}} \sum_{n \in p_i}\mathcal{H}(||p_i^n - \psi(p_i^n,m_i)||^2, \delta)~,
    \label{eq-7a}
\end{equation}
\begin{equation}
    \mathcal{L}^i_{backward} = \frac{1}{N_{m_i}} \sum_{n \in m_i}\mathcal{H}(||m_i^n - \psi(m_i^n,p_i)||^2, \delta)~,
    \label{eq-7b}
\end{equation}
\end{subequations}

\begin{equation}
    \mathcal{L}_{icp} = \frac{1}{N_L}\sum_{i\in L}(\mathcal{L}^i_{forward} + \mathcal{L}^i_{backward})~,
    \label{eq-8}
\end{equation}
where $N_{p_i}$, $N_{m_i}$ are the number of pixels in $p_i$ and $m_i$, $N_L$ is the number of matching objects, and $\psi(\cdot)$ outputs the closest peer of a pixel. Our dual-path 2D ICP loss in Eq.~\ref{eq-8} can make full use of object information to align not only centers of objects but also their entire 2D shapes. The importance of bidirectional design lies in eliminating scale ambiguity arising from $\mathcal{L}_{forward}$ or $\mathcal{L}_{backward}$ alone. Benefiting from this loss, we can achieve stable pose optimization. 

\section{Experiments}
\label{sec:experiments}
In this section, we describe our experimental setup and validate that our system can achieve significant improvements in relocalization performance. We evaluate our system in single-room (Sec.~\ref{sec4-1}) and multi-room/floor (Sec.~\ref{sec4-2}) scenes, and perform extended analysis (Sec.~\ref{sec4-3+}). We also evaluate its object matching performance (Sec.~\ref{sec4-3}), map size (Sec.~\ref{sec4-4}), and robustness (Sec.~\ref{sec:lighting}). The effectiveness of our designs is confirmed in Sec.~\ref{sec4-5}. The best results are highlighted \textbf{in bold} in tables.

\noindent\textbf{Datasets.} In the single-room case, we utilize two real-world RGB-D indoor datasets: ScanNet~\cite{scannet} and ScanNet++~\cite{scannet++}. They contain rich object categories and diverse scenes without temporal changes, but only provide sequential frames with high visual overlap. To ensure a notable view difference between mapping and query images, we split each sequence into two non-overlapping subsets. For example, in ScanNet, mapping frames are sampled from frame 0 at a stride of 20, forming \{0, 20, 40, …\}, while query frames start from frame 10 with the same stride, forming \{10, 30, 50, …\}. In the multi-room/floor case, we generate a larger-scale Synthetic dataset, enabling free synthesis of mapping and query frames with significant viewpoint differences.

\noindent\textbf{Implementation Details.} We run our system on a desktop equipped with an NVIDIA RTX 4090 GPU. We set the learning rate of $\{q, T\}$ to \{0.025, 0.025\} in the refined pose optimization. During extracting 3D landmark descriptors, we select $k$=5 segmentation patches. In 2D ICP loss, we use the hyperparameter $\delta=10$ in the Huber kernel. The path length $\eta$ is set to $\eta=1$ in the node search to extract 3D subgraphs from the global scene graph. Our system selects the official ViT-L/14@336px CLIP model to embed 768-dimension open-vocabulary features.

\noindent\textbf{Metrics.} A relocalization system is most concerned about its success rate and pose accuracy. We quantitatively evaluate these two aspects using different metrics. With respect to success rate, we count the percentage of correctly relocalized query images within given translation thresholds: 50cm and 25cm, \ie, $Recall[\%]$ at 50cm and 25cm. As for accuracy, we calculated the mean translation error ($MTE[cm]$) and mean rotation error ($MRE[^\circ]$) for those query images within $Recall@\textit{50cm}$ and $Recall@\textit{25cm}$, respectively.

\noindent\textbf{Baselines.} As a novel paradigm for camera relocalization, OpenReLoc should be compared with some object-level 6-DOF relocalization methods. However, limited open-source work is available in this emerging research field. Therefore, our main comparison is to GoReloc~\cite{goreloc}, an open-source and SOTA object-level baseline, which shares the most relevant problem formulation with ours. Additionally, we also include several low-level vision methods~\cite{coordinet, ms-transformer, mur2017orb,pixloc} as additional baselines for completeness.

\begin{table}[t]
    \centering
    \LARGE
    \renewcommand{\arraystretch}{1.7}
    \caption{\textbf{Recall and Accuracy on ScanNet.} Each cell shows @50cm\,/\,@25cm.}
    \resizebox{0.49\textwidth}{!}{
    \begin{tabular}{cccccccccc}
        \toprule
        Method & Metric & 0568 & 0101 & 0673 & 0108 & 0166 & 0378 & 0092 & 0603 \\
        \midrule
        \multirow{3}{*}{CoordiNet~\cite{coordinet}} & 
        Recall[\%]$\uparrow$ & 36\,/\,6 & 17\,/\,3 & 32\,/\,8 & 18\,/\,4 & 61\,/\,21 & 44\,/\,17 & 38\,/\,6 & 33\,/\,9 \\ &
        MTE[m]$\downarrow$ & 0.34\,/\,0.21 & 0.35\,/\,0.21 & 0.32\,/\,0.18 & 0.36\,/\,0.23 & 0.31\,/\,0.17 & 0.32\,/\,0.2 & 0.35\,/\,0.18 & 0.32\,/\,0.17 \\ & 
        MRE[$^\circ$]$\downarrow$ & 13.6\,/\,19.2 & 17.8\,/\,8.2 & 10.5\,/\,9.0 & 21.6\,/\,31.1 & 11.3\,/\,11.4 & 14.8\,/\,14.1 & 14.5\,/\,8.2 & 11.3\,/\,12.5 \\
        \midrule
        \multirow{3}{*}{MS-Transformer~\cite{ms-transformer}} & 
        Recall[\%]$\uparrow$ & 76\,/\,32 & 37\,/\,14 & 14\,/\,4 & 82\,/\,57 & 52\,/\,33 & 41\,/\,21 & 45\,/\,19 & 40\,/\,9 \\ & 
        MTE[m]$\downarrow$ & 0.28\,/\,0.16 & 0.31\,/\,0.18 & 0.32\,/\,0.23 & 0.21\,/\,0.15 & 0.23\,/\,0.17 & 0.26\,/\,0.16 & 0.25\,/\,0.14 & 0.32\,/\,0.16 \\ & 
        MRE[$^\circ$]$\downarrow$ & 23.2\,/\,18.6 & 43.4\,/\,35.8 & 46.8\,/\,9.4 & 41.7\,/\,31.5 & 27.4\,/\,21.2 & 33.4\,/\,34.2 & 15.3\,/\,12.7 & 25.5\,/\,22.0 \\
        \midrule
        \multirow{3}{*}{GoReloc~\cite{goreloc}} & 
        Recall[\%]$\uparrow$ & 8\,/\,5 & 17\,/\,17 & 22\,/\,10 & 25\,/\,12 & 16\,/\,7 & 12\,/\,3 & 9\,/\,6 & 14\,/\,5 \\ & 
        MTE[m]$\downarrow$ & 0.23\,/\,0.14 & 0.21\,/\,0.21 & 0.29\,/\,0.14 & 0.26\,/\,0.12 & 0.23\,/\,0.13 & 0.30\,/\,0.16 & 0.23\,/\,0.14 & 0.29\,/\,0.14 \\ & 
        MRE[$^\circ$]$\downarrow$ & 4.6\,/\,\textbf{2.1} & 4.9\,/\,4.9 & 10.2\,/\,4.5 & 4.5\,/\,3.7 & 7.0\,/\,4.9 & 9.5\,/\,5.2 & 8.6\,/\,7.4 & 26.8\,/\,6.1 \\
        \midrule
        \multirow{3}{*}{\textbf{Ours}} & 
        Recall[\%]$\uparrow$ & \textbf{79}\,/\,\textbf{58} & \textbf{68}\,/\,\textbf{45} & \textbf{64}\,/\,\textbf{51} & \textbf{89}\,/\,\textbf{73} & \textbf{72}\,/\,\textbf{52} & \textbf{83}\,/\,\textbf{74} & \textbf{69}\,/\,\textbf{50} & \textbf{65}\,/\,\textbf{52} \\ & 
        MTE[m]$\downarrow$ & \textbf{0.18}\,/\,\textbf{0.13} & \textbf{0.2}\,/\,\textbf{0.14} & \textbf{0.17}\,/\,\textbf{0.12} & \textbf{0.16}\,/\,\textbf{0.11} & \textbf{0.18}\,/\,\textbf{0.12} & \textbf{0.12}\,/\,\textbf{0.09} & \textbf{0.19}\,/\,\textbf{0.12} & \textbf{0.15}\,/\,\textbf{0.10} \\ &
        MRE[$^\circ$]$\downarrow$ & \textbf{4.0}\,/\,2.9 & \textbf{4.7}\,/\,\textbf{3.6} & \textbf{4.6}\,/\,\textbf{3.4} & \textbf{3.9}\,/\,\textbf{2.8} & \textbf{6.0}\,/\,\textbf{4.0} & \textbf{3.6}\,/\,\textbf{2.7} & \textbf{6.0}\,/\,\textbf{4.0} & \textbf{5.7}\,/\,\textbf{3.8} \\
        \bottomrule
    \end{tabular}
    }
    \vspace{-0.25em}
    \label{tab:scannet}
\end{table}

\begin{table}[t]
    \centering
    \renewcommand{\arraystretch}{1.7}
    \LARGE
    \caption{\textbf{Recall and Accuracy on ScanNet++.} Each cell shows @50cm\,/\,@25cm.}
    \resizebox{0.49\textwidth}{!}{
    \begin{tabular}{cccccccccc}
        \toprule
        Method & Metric & 0a7cc & 0a184 & 0d2ee & 7e094 & 7f4d1 & 25f3b & 8890d & a08d9 \\
        \midrule
        \multirow{3}{*}{CoordiNet~\cite{coordinet}} & 
        Recall[\%]$\uparrow$ & 35\,/\,10 & 39\,/\,7 & 32\,/\,5 & 62\,/\,21 & 36\,/\,9 & 54\,/\,15 & 64\,/\,13 & 35\,/\,11 \\ &
        MTE[m]$\downarrow$ & 0.31\,/\,0.18 & 0.35\,/\,0.16 & 0.37\,/\,0.20 & 0.33\,/\,0.21 & 0.33\,/\,0.16 & 0.32\,/\,0.19 & 0.32\,/\,0.15 & 0.34\,/\,0.19 \\ & 
        MRE[$^\circ$]$\downarrow$ & 13.3\,/\,14.0 & 13.4\,/\,12.2 & 26.1\,/\,22.5 & 6.2\,/\,5.9 & 11.3\,/\,13.1 & 10.1\,/\,8.2 & 6.4\,/\,6.5 & 11.8\,/\,13.7 \\
        \midrule
        \multirow{3}{*}{MS-Transformer~\cite{ms-transformer}} & 
        Recall[\%]$\uparrow$ & 68\,/\,41 & 72\,/\,32 & 32\,/\,11 & 60\,/\,35 & 57\,/\,18 & 60\,/\,40 & 66\,/\,27 & 51\,/\,11 \\ & 
        MTE[m]$\downarrow$ & 0.25\,/\,0.17 & 0.26\,/\,0.16 & 0.31\,/\,0.17 & 0.22\,/\,0.13 & 0.29\,/\,0.16 & 0.22\,/\,0.16 & 0.28\,/\,0.17 & 0.32\,/\,0.18 \\ & 
        MRE[$^\circ$]$\downarrow$ & 14.2\,/\,12.2 & 26.8\,/\,25.3 & 39.8\,/\,26.0 & 38.2\,/\,13.1 & 25.4\,/\,23.4 & 22.8\,/\,20.2 & 26.3\,/\,17.3 & 18.7\,/\,12.1 \\
        \midrule
        \multirow{3}{*}{\textbf{Ours}} & 
        Recall[\%]$\uparrow$ & \textbf{70}\,/\,\textbf{60} & \textbf{74}\,/\,\textbf{59} & \textbf{62}\,/\,\textbf{54} & \textbf{92}\,/\,\textbf{80} & \textbf{75}\,/\,\textbf{68} & \textbf{77}\,/\,\textbf{65} & \textbf{81}\,/\,\textbf{70} & \textbf{70}\,/\,\textbf{54} \\ & 
        MTE[m]$\downarrow$ & \textbf{0.11}\,/\,\textbf{0.07} & \textbf{0.12}\,/\,\textbf{0.07} & \textbf{0.13}\,/\,\textbf{0.10} & \textbf{0.11}\,/\,\textbf{0.08} & \textbf{0.09}\,/\,\textbf{0.06} & \textbf{0.13}\,/\,\textbf{0.09} & \textbf{0.09}\,/\,\textbf{0.06} & \textbf{0.16}\,/\,\textbf{0.09} \\ &
        MRE[$^\circ$]$\downarrow$ & \textbf{4.0}\,/\,\textbf{2.7} & \textbf{4.1}\,/\,\textbf{2.5} & \textbf{4.2}\,/\,\textbf{3.7} & \textbf{3.7}\,/\,\textbf{2.4} & \textbf{3.1}\,/\,\textbf{2.6} & \textbf{7.1}\,/\,\textbf{5.1} & \textbf{3.7}\,/\,\textbf{2.0} & \textbf{8.0}\,/\,\textbf{4.0} \\
        \bottomrule
    \end{tabular}
    }
    \vspace{-0.7em}
    \label{tab:scannetpp}
\end{table}

\vspace{-0.6em}
\subsection{Single-room Relocalization}
\label{sec4-1}

\noindent\textbf{Evaluation on ScanNet~\cite{scannet}.} We report relocalization results of our system across 8 ScanNet scenes in Tab.~\ref{tab:scannet}. Despite incomplete and noisy RGB-D observations, multi-modal landmark association and dual-path ICP pose refinement promote a stable convergence of our system to a correct solution. It is clear that our system surpasses all other baselines with a notable margin in both success rate and accuracy. Against GoReloc~\cite{goreloc}, our method substantially increases the success rate by around \textbf{5$\sim$10} times.

\noindent\textbf{Evaluation on ScanNet++~\cite{scannet++}.} As shown in Tab.~\ref{tab:scannetpp}, we also evaluate the relocalization performance of our system on ScanNet++. In GoReloc~\cite{goreloc}, mapping heavily depends on YOLOv8~\cite{yolo} and ORB-SLAM2~\cite{mur2017orb} features, while severe weak-texture conditions in ScanNet++ hinder its normal operation. Therefore, we only compare other methods to ours on this dataset. Similarly, our system still achieves better relocalization performance, and the high-quality sensor data in ScanNet++ further facilitates higher pose accuracy (\textit{MTE} and \textit{MRE}) of our system.

These ScanNet and ScanNet++ scenes are captured from the real world, where rich object diversity falls beyond the closed-vocabulary scope of GoReloc. As a result, GoReloc fails to identify valid matching objects in many observations. This factor explains GoReloc's poor success rate. In addition, GoReloc does not incorporate a dedicated optimization loss. This means that even when GoReloc successfully identifies matching objects, it still lacks stable optimization guidance, making it prone to drift and suboptimal convergence. This factor is primarily responsible for GoReloc's poor accuracy. Experiments on these two datasets illustrate the capability of our system in handling complex real-world scenes, boosting the practicality of object-level camera relocalization.

\vspace{-0.5em}
\subsection{Multi-room Relocalization} 
\label{sec4-2}

\begin{figure}[!t]
    \centering
    \includegraphics[width = \linewidth]{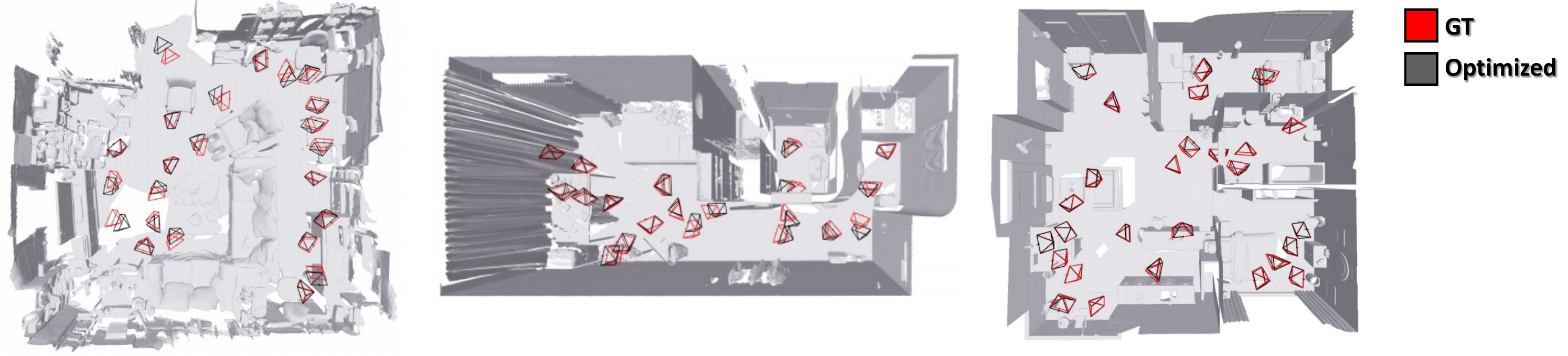}
    \caption{\textbf{Qualitative visualization.} We qualitatively show relocalization poses and their ground truth on various scenes.}
    \label{fig:exp-results}
    \vspace{-0.4em}
\end{figure}

With available assets in the HSSD~\cite{khanna2024habitat}, we assembled challenging indoor scenes, consisting of multi-room (Scenes 1–6) and multi-floor (Scenes 7–8) cases, to evaluate generalization and scalability. As shown in Tab.~\ref{tab:synthetic}, our system exhibits robustness in this large-scale setting, excelling all baselines. GoReloc~\cite{goreloc} and CoordiNet~\cite{coordinet} operate poorly in $Recall@\textit{50cm}$ and fail entirely in $Recall@\textit{25cm}$, prompting us to introduce a more lenient $Recall@\textit{100cm}$ metric. Our system's scalability is attributed to the coarse-to-fine mechanism utilizing object-oriented reference frames and DIOU-based retrieval, whereas the absence of coarse priors in GoReloc~\cite{goreloc} severely hinders it from scaling to large-scale scenes. Relocalization results on different datasets are visualized in Fig.~\ref{fig:exp-results}.

\begin{table}[t]
    \centering
    \LARGE
    \renewcommand{\arraystretch}{1.7}
    \caption{\textbf{Recall and Accuracy on Synthetic.} Each cell shows @50cm\,/\,@25cm for MS-Transformer and Ours, and @50cm\,/\,@100cm for CoordiNet and GoReloc.}
    \resizebox{0.49\textwidth}{!}{
    \begin{tabular}{cccccccccc}
        \toprule
        Method & Metric & Sc-1 & Sc-2 & Sc-3 & Sc-4 & Sc-5 & Sc-6 & Sc-7 & Sc-8 \\
        \midrule
        \multirow{3}{*}{\makecell{CoordiNet~\cite{coordinet}\\(@50\,/\,@100cm)}} & 
        Recall[\%]$\uparrow$ & 7\,/\,19 & 13\,/\,46 & 7\,/\,29 & 8\,/\,32 & 10\,/\,33 & 15\,/\,40 & 3\,/\,24 & 9\,/\,45 \\ &
        MTE[m]$\downarrow$ & 0.31\,/\,0.58 & 0.32\,/\,0.62 & 0.43\,/\,0.70 & 0.27\,/\,0.61 & 0.33\,/\,0.60 & 0.38\,/\,0.63 & 0.27\,/\,0.68 & 0.37\,/\,0.68 \\ & 
        MRE[$^\circ$]$\downarrow$ & 14.1\,/\,29.7 & 13.1\,/\,15.9 & 33.7\,/\,24.8 & 17.4\,/\,20.6 & 18.9\,/\,18.3 & 15.3\,/\,17.9 & 11.9\,/\,15.6 & 11.7\,/\,13.9 \\
        \midrule
        \multirow{3}{*}{\makecell{GoReloc~\cite{goreloc}\\(@50\,/\,@100cm)}} & 
        Recall[\%]$\uparrow$ & 7\,/\,11 & 8\,/\,17 & 0\,/\,10 & 0\,/\,25 & 21\,/\,21 & 29\,/\,35 & 0\,/\,0 & 0\,/\,0 \\ & 
        MTE[m]$\downarrow$ & 0.41\,/\,0.58 & 0.38\,/\,0.54 & -\,/\,0.59 & -\,/\,0.56 & 0.21\,/\,0.21 & 0.35\,/\,0.48 & -\,/\,- & -\,/\,- \\ & 
        MRE[$^\circ$]$\downarrow$ & 10.6\,/\,11.5 & 14.6\,/\,18.0 & -\,/\,12.5 & -\,/\,33.2 & 5.8\,/\,5.8 & 12.7\,/\,18.2 & -\,/\,- & -\,/\,- \\
        \midrule
        \multirow{3}{*}{\makecell{MS-Transformer~\cite{ms-transformer}\\(@50\,/\,@25cm)}} & 
        Recall[\%]$\uparrow$ & 13\,/\,4 & 30\,/\,8 & 34\,/\,7 & 53\,/\,20 & 28\,/\,13 & 35\,/\,4 & 16\,/\,5 & 27\,/\,9 \\ & 
        MTE[m]$\downarrow$ & 0.33\,/\,0.16 & 0.33\,/\,0.21 & 0.34\,/\,0.17 & 0.31\,/\,0.18 & 0.28\,/\,0.14 & 0.35\,/\,0.16 & 0.33\,/\,0.16 & 0.31\,/\,0.17 \\ & 
        MRE[$^\circ$]$\downarrow$ & 4.0\,/\,4.9 & 6.2\,/\,6.8 & \textbf{4.5}\,/\,3.6 & 7.4\,/\,3.5 & 5.6\,/\,4.7 & 7.8\,/\,14.1 & 4.1\,/\,3.6 & 7.2\,/\,3.8 \\
        \midrule
        \multirow{3}{*}{\makecell{\textbf{Ours}\\(@50\,/\,@25cm)}} & 
        Recall[\%]$\uparrow$ & \textbf{86}\,/\,\textbf{73} & \textbf{88}\,/\,\textbf{83} & \textbf{89}\,/\,\textbf{75} & \textbf{90}\,/\,\textbf{69} & \textbf{91}\,/\,\textbf{86} & \textbf{86}\,/\,\textbf{74} & \textbf{79}\,/\,\textbf{71} & \textbf{83}\,/\,\textbf{75} \\ & 
        MTE[m]$\downarrow$ & \textbf{0.12}\,/\,\textbf{0.07} & \textbf{0.07}\,/\,\textbf{0.05} & \textbf{0.11}\,/\,\textbf{0.06} & \textbf{0.14}\,/\,\textbf{0.07} & \textbf{0.09}\,/\,\textbf{0.06} & \textbf{0.09}\,/\,\textbf{0.05} & \textbf{0.1}\,/\,\textbf{0.06} & \textbf{0.09}\,/\,\textbf{0.06} \\ &
        MRE[$^\circ$]$\downarrow$ & \textbf{3.8}\,/\,\textbf{1.8} & \textbf{2.4}\,/\,\textbf{1.9} & 4.7\,/\,\textbf{2.3} & \textbf{5.4}\,/\,\textbf{2.6} & \textbf{4.1}\,/\,\textbf{2.7} & \textbf{4.2}\,/\,\textbf{2.0} & \textbf{3.4}\,/\,\textbf{2.2} & \textbf{3.6}\,/\,\textbf{2.5} \\
        \bottomrule
    \end{tabular}
    }
    \begin{tablenotes}
    \footnotesize
    \item[*] `\textbf{-}' denotes failure cases.
    \end{tablenotes}
    \vspace{-0.7em}
    \label{tab:synthetic}
\end{table}

\vspace{-0.5em}
\subsection{Extended Analysis}
\label{sec4-3+}

\noindent\textbf{Comparison with ORB-SLAM2 and PixLoc.} For a more thorough evaluation, we also compare our method with some low-level vision methods that employ a structural pipeline, ORB-SLAM2~\cite{mur2017orb} and PixLoc~\cite{pixloc}, on the Synthetic dataset with the largest scene scale and viewpoint differences. Average results across scenes in Tab.~\ref{review1:pixloc and orbslam2} show that ORB-SLAM2 experienced failure, succeeding on very few frames, despite achieving better accuracy. This stems from the limited robustness of hand-crafted ORB features. Compared to PixLoc, our method achieves higher recall owing to open-vocabulary capability and environmental robustness. In terms of accuracy, PixLoc benefits from fine-grained feature point correspondences, whereas our coarse-grained object correspondences yield slightly lower accuracy. Nevertheless, our map size is one-tenth that of PixLoc, offering a much more compact representation.

\begin{table}[t]
    \centering
    \Large
    \caption{\textbf{Comparison with PixLoc and ORB-SLAM2.} This table shows average metrics over multi-room/floor scenes of the Synthetic dataset.}
    \resizebox{0.48\textwidth}{!}{
     \begin{tabular}{cccccccc}
        \toprule
        \multirow{2}{*}{\textbf{Method}} & \multicolumn{3}{c}{\textbf{@50cm}} & \multicolumn{3}{c}{\textbf{@25cm}} & \textbf{Map Size} \\
        \cmidrule(lr){2-4} \cmidrule(lr){5-7} \cmidrule(lr){8-8}
        & Recall[\%]$\uparrow$ & MTE[m]$\downarrow$ & MRE[$^\circ$]$\downarrow$ & Recall[\%]$\uparrow$ & MTE[m]$\downarrow$ & MRE[$^\circ$]$\downarrow$ & [MB]$\downarrow$ \\
        \midrule
        \textbf{Ours} & \textbf{87}  & 0.10  &  3.9 & \textbf{81}  &  0.06 & 2.2 & \textbf{24.5} \\
        \midrule
        \textbf{PixLoc} & 82  & 0.07  &  2.7 &  80 &  0.04 &  1.6 & 273.8 \\
        \midrule
        \textbf{ORB-SLAM2} & 17  & \textbf{0.02}  & \textbf{1.0}  & 17  & \textbf{0.02}  & \textbf{1.0}  & 262.4 \\
        \bottomrule
    \end{tabular}
    }
    \vspace{-0.3em}
    \label{review1:pixloc and orbslam2}
\end{table}

\noindent\textbf{Comparison with GoReloc on TUM-RGBD.} For a more intuitive comparison with GoReloc, we conduct an experiment on the TUM RGB-D~\cite{tumrgbd} dataset under the same experimental setup as in the GoReloc paper. In Tab.~\ref{review1:tumrgbd}, it can be seen that our method can still outperform GoReloc in both success rate and accuracy.

\begin{table}[t]
\renewcommand{\arraystretch}{0.8}
	\centering
	\caption{\textbf{Evaluation on TUM RGB-D.} Comparison with GoReloc under the same experimental setup.}  
    \resizebox{0.38\textwidth}{!}{
	\begin{tabular}{cccccc} 
    \toprule 
    \multirow{2}{*}{\textbf{Dataset}} & \multirow{2}{*}{\textbf{Method}} & \multicolumn{2}{c}{\textbf{Success Rate} ($\%$) $\uparrow$} & \multicolumn{2}{c}{\textbf{TE}(m) $\downarrow$}  
 \\ 
 \cmidrule(lr){3-4}\cmidrule(lr){5-6} 
    & & @2m & @5m & $10\%$ & $20\%$ 
 \\ 
  \midrule
 \multirow{2}{*}{\textbf{TUM RGB-D}} 
	& \textbf{GoReloc}  & 64.87 & 96.11 & 0.73  & 0.90\\
    \cmidrule(lr){2-6}
    & \textbf{Ours}   & \textbf{89.42} & \textbf{98.78} & \textbf{0.13}  & \textbf{0.18} 
   \\ 
		\bottomrule
	\end{tabular} 
    }
    \vspace{-0.3em}
 \label{review1:tumrgbd}
\end{table}

\begin{table}[!t]
    \centering
    \renewcommand{\arraystretch}{1.3}
    \Large
    \caption{\textbf{Efficiency Analysis.} Per-frame runtime and a detailed breakdown of ours at different stages.}
    \Large
    \resizebox{0.48\textwidth}{!}{
    \begin{tabular}{ccccccc}
    \toprule
    \multirow{2}{*}{  \textbf{Metric}} & \multirow{2}{*}{ \textbf{PixLoc}} & \multicolumn{5}{c}{  \textbf{Ours}} \\
    \cmidrule(lr){3-7}
    & & \textbf{Object Detection} & \textbf{GPT Analysis} & \textbf{CLIP Encoding} & \textbf{Coarse-to-fine Pose} & \textbf{Total} \\
    \midrule
    \textbf{Runtime} & $\approx${\LARGE 4.5s} & $\approx${\LARGE 0.3s} & $\approx${\LARGE 4.1s} & $\approx${\LARGE 0.2s} & $\approx${\LARGE 0.5s} & $\approx${\LARGE 5.1s} \\
    \bottomrule
    \end{tabular}
    }
    \vspace{-0.1em}
    \label{review2:efficiency}
\end{table}

\begin{figure}[!t]
    \centering
    \includegraphics[width = \linewidth]{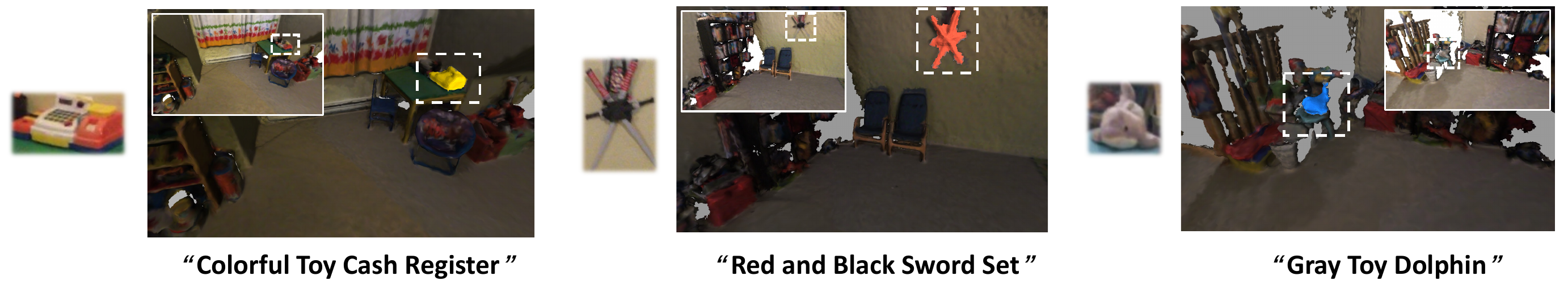}
    \caption{\textbf{Open-vocabulary Object Matching.} Open-vocabulary object-level mapping allows us to recognize diverse objects.}
    \label{fig:ov-matching}
    \vspace{-1.3em}
\end{figure}

\noindent\textbf{Efficiency Analysis.} We compare per-frame runtime with PixLoc and report our runtime breakdown in Tab.~\ref{review2:efficiency}. Both methods achieve comparable efficiency. The relocalization module typically serves as an initial pose provider or a fallback after tracking failures. As such, it does not demand strict real-time performance but places greater emphasis on success rate and accuracy. Unlike GoReloc, which excessively prioritizes high efficiency at the cost of system performance, ours and PixLoc strike a more balanced trade-off among success rate, accuracy, and efficiency. Notably, in our system, GPT analysis is a major efficiency bottleneck, accounting for about 80\% of the per-frame runtime due to the latency of online API calls, as the closed-source GPT does not support local deployment.

\vspace{-0.5em}
\subsection{Open-vocbulary Object Matching}
\label{sec4-3}
Notably, OpenReLoc can recognize class-agnostic objects in an open-vocabulary manner, which essentially sets ours apart from GoReloc~\cite{goreloc}. Scenes used in our experiments all exhibit a long-tail object distribution. Such a distribution falls beyond the scope of closed-vocabulary methods, leading to their failure. In contrast, our open-vocabulary capability can still achieve effective matching on these scenes, and we also qualitatively present several successfully matched long-tail examples along with their language descriptions in Fig.~\ref{fig:ov-matching}.

\vspace{-0.5em}
\subsection{Map Size Analysis} 
\label{sec4-4}
We also report the map memory consumption of different baselines on the ScanNet ‘0568’ scene in Tab.~\ref{tab:memory}, where object-level methods (GoReloc~\cite{goreloc} and Ours) can construct a more compact map compared to low-level vision methods. Furthermore, compared to GoReloc~\cite{goreloc}, removing object color and category likelihood saves an additional \textbf{80\%} of memory consumption in our system.

\begin{table}[t]
  \centering
  \footnotesize
  \renewcommand{\arraystretch}{1.1}
  \caption{\textbf{Map Size.} Map memory consumption on the single-room ScanNet ‘0568’ scene.}
  \resizebox{0.49\textwidth}{!}{
      \begin{tabular}{ccccc}
        \toprule
        Metric & CoordiNet~\cite{coordinet} & MS-Transformer~\cite{ms-transformer} & GoReloc~\cite{goreloc} & \textbf{Ours} \\
        \midrule
        Map Size & 71.4~MB & 63.1~MB & 17.2~MB & \textbf{3.5~MB}\\
        \bottomrule
      \end{tabular}
      }
    \vspace{-0.25em}
  \label{tab:memory}
\end{table}

\begin{table}[!t]
    \centering
    \caption{\textbf{Evaluation under Lighting Variations.} Quantitative impact of lighting variations on recall and accuracy.}
    \renewcommand{\arraystretch}{0.7}
    \resizebox{0.48\textwidth}{!}{
\begin{tabular}{ccccccc}
    \toprule
    \multirow{2}{*}{Setting}  & \multicolumn{3}{c}{\textbf{\textit{@50cm}}} & \multicolumn{3}{c}{\textbf{\textit{@25cm}}} \\
    \cmidrule(lr){2-4} \cmidrule(lr){5-7}
    & Recall[\%]$\uparrow$ & MTE[m]$\downarrow$ & MRE[$^\circ$]$\downarrow$ & Recall[\%]$\uparrow$ & MTE[m]$\downarrow$ & MRE[$^\circ$]$\downarrow$\\
    \midrule
    {\textbf{Original}} & \textbf{81}  & \textbf{0.09}  & \textbf{3.7}  & \textbf{70}  & \textbf{0.06}  & \textbf{2.0}  \\
     \midrule
    {\large\ding{172}}  & 79  & 0.12  & 5.3  & \textbf{70}  & 0.07  & 2.4  \\
    \midrule
    {\large\ding{173}}  & 75  & 0.13  & 5.6  & 62 & \textbf{0.06} & 2.9 \\
    \midrule
    {\large\ding{174}}  & 70  & 0.15  & 5.7  & 56  & 0.09  & 2.9  \\
    \midrule
    {\large\ding{175}}  & 66  & 0.16  & 6.0  & 55  & 0.10  & 3.1  \\
    \bottomrule
\end{tabular}
    }
    \label{review1:lighting eval}
\end{table}

\begin{figure}[!t]
    \centering
    \includegraphics[width = \linewidth]{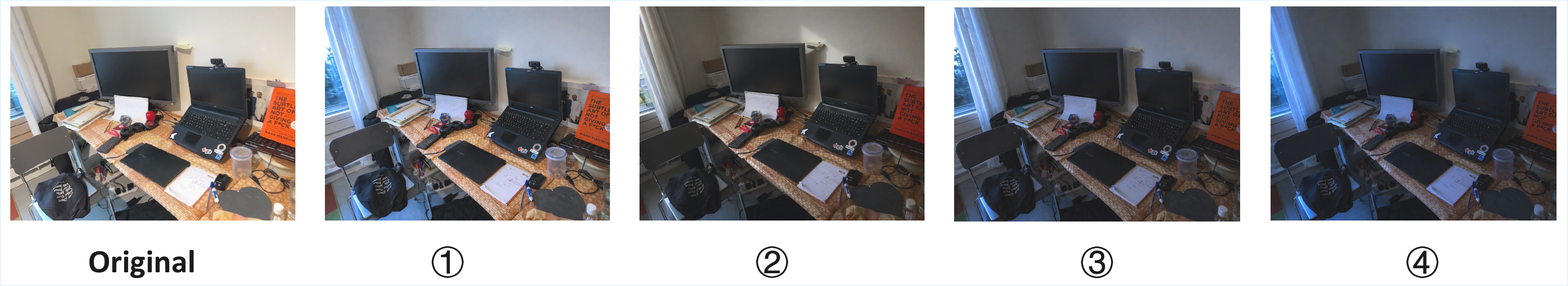}
    \caption{\textbf{Lighting Variation.} We display the scene appearance under progressive illumination decay.}
    \label{fig:lighting}
    \vspace{-0.5em}
\end{figure}

\begin{figure}[!t]
    \centering
    \includegraphics[width = 1.0\linewidth]{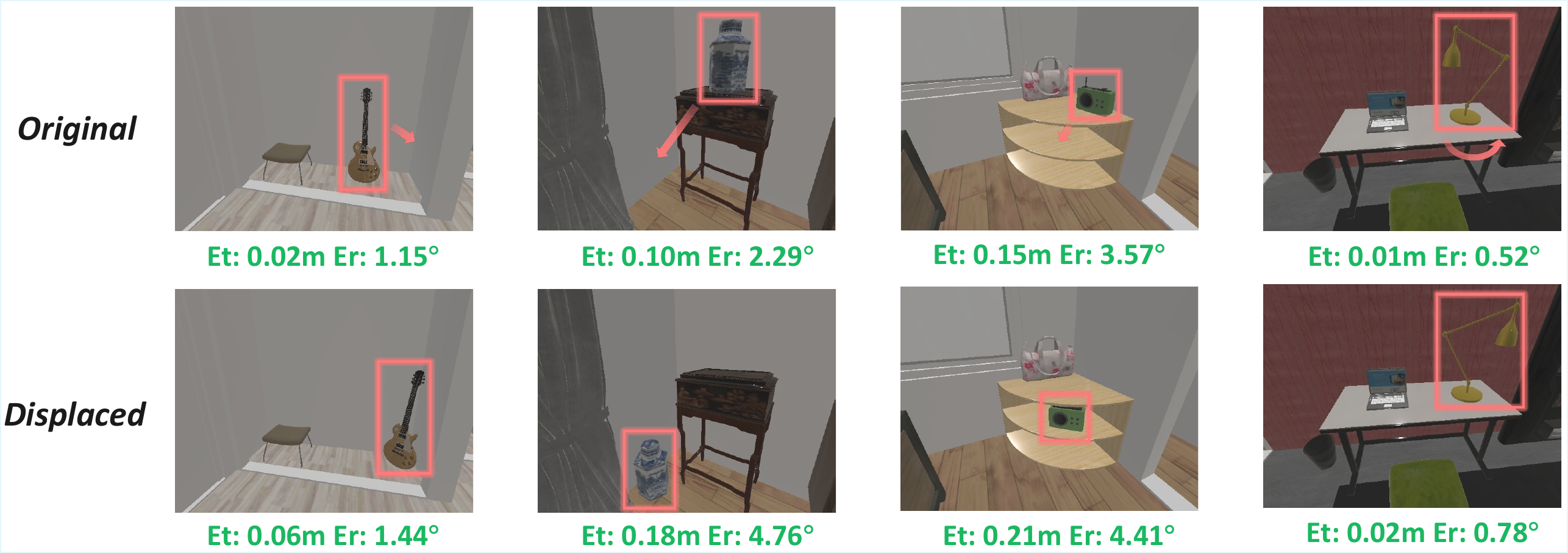}
    \caption{\textbf{Object Displacement.} We displaced several objects in synthetic scenes and evaluated quantitative metrics. }
    \label{fig:displaced}
    \vspace{-1.3em}
\end{figure}

\begin{table}[t]
\centering
\footnotesize
\renewcommand{\arraystretch}{1.7}
\LARGE
\caption{\textbf{Ablation Study.} Ablations of main module designs on different datasets. Each cell shows @50cm\,/\,@25cm.}
\resizebox{0.48\textwidth}{!}{
    \begin{tabular}{
        l
        ccc @{\hspace{20pt}}
        ccc @{\hspace{20pt}}
        ccc
    }
    \toprule
    \textbf{Setting}
    & \multicolumn{3}{c@{\hspace{20pt}}}{Recall[\%]$\uparrow$}
    & \multicolumn{3}{c@{\hspace{20pt}}}{MTE[m]$\downarrow$}
    & \multicolumn{3}{c}{MRE[$^{\circ}$]$\downarrow$} \\
    \cmidrule(lr){2-4}\cmidrule(lr){5-7}\cmidrule(lr){8-10}
    & 0568 & 0a7cc & scene1
    & 0568 & 0a7cc & scene1
    & 0568 & 0a7cc & scene1 \\
    \midrule
    \textbf{\#1}~~w/o Refine Stage
    & 62/46 & 45/41 & 6/\textbf{-}
    & 0.25/0.18 & 0.15/0.09 & 0.43/\textbf{-}
    & 8.5/6.1 & 11.0/11 & 29.7/\textbf{-}
    \\
    \midrule
    \textbf{\#2}~~w/o Coarse Stage
    & 27/11 & \textbf{-}/\textbf{-} & 29/20
    & 0.29/0.15 & \textbf{-}/\textbf{-} & 0.17/0.09
    & 8.3/3.3 & \textbf{-}/\textbf{-} & 5.0/2.9
    \\
    \midrule
    \textbf{\#3}~~w/o Scene Graph
    & 64/41 & 64/56 & 78/70
    & 0.21/0.14 & \textbf{0.11}/0.08 & \textbf{0.12}/0.08
    & 4.6/3.1 & 4.6/4.0 & \textbf{3.5}/2.1
    \\
    \midrule
    \textbf{\#4}~~w/o Language Modality
    & 66/49 & 52/44 & 76/62
    & \textbf{0.18}/\textbf{0.13} & 0.12/\textbf{0.07} & 0.13/\textbf{0.07}
    & 4.1/\textbf{2.9} & 4.4/2.8 & 4.0/\textbf{1.8}
    \\
    \midrule
    \textbf{\#5}~~w/o DIOU-based Retrieval
    & 74/50 & 55/48 & 75/53
    & 0.19/0.16 & 0.13/0.09 & 0.16/0.11
    & 4.2/3.5 & 4.5/3.4 & 4.4/2.8
    \\
    \midrule
    \textbf{\#6}~~w/o Invalid Object Filtering
    & 77/55 & 64/53 & 75/60
    & \textbf{0.18}/\textbf{0.13} & 0.12/0.08 & 0.14/0.09
    & 4.2/\textbf{2.9} & 4.3/3.3 & 3.8/2.5
    \\
    \midrule
    \textbf{Ours Full}
    & \textbf{79}/\textbf{58} & \textbf{70}/\textbf{60} & \textbf{86}/\textbf{73}
    & \textbf{0.18}/\textbf{0.13} & \textbf{0.11}/\textbf{0.07} & \textbf{0.12}/\textbf{0.07}
    & \textbf{4.0}/\textbf{2.9} & \textbf{4.0}/\textbf{2.7} & 3.8/\textbf{1.8}
    \\[-0.4ex]
    \bottomrule
    \end{tabular}
}
\begin{tablenotes}
    \footnotesize
    \item[*] `\textbf{-}' denotes failure cases.
\end{tablenotes}
\vspace{-0.7em}
\label{tab:ablation-modality}
\end{table}

\vspace{-0.7em}
\subsection{Robustness Analysis} 
\label{sec:lighting}
Object-level approaches are inherently robust to environmental changes such as lighting variations and object displacements~\cite{zins2022oa,yang2019cubeslam}. Such robustness is rooted in the stability of semantic information. As shown in Fig.~\ref{fig:lighting} and Tab.~\ref{review1:lighting eval}, an advanced VLM (GPT-Image) is used to simulate lighting variations, and our method experiences moderate performance degradation under drastic lighting attenuation. We further simulate object displacements in synthetic scenes, like human-induced rearrangements (Fig.~\ref{fig:displaced}). At each optimization step, we first calculate the ICP loss value for every object, and then, following prior work~\cite{zhu2022nice}, we mask out any object whose ICP loss exceeds 5× the median loss before backpropagation. This strategy adaptively filters out moving items during optimization. Quantitative metrics in Fig.~\ref{fig:displaced} show that our approach remains effective under moderate object displacements. 

\vspace{-0.5em}
\subsection{Ablation Study}
\label{sec4-5}
To verify the rationality of our main module designs, we conduct ablation studies on different datasets in Tab.~\ref{tab:ablation-modality}.

\noindent\textbf{Coarse-to-fine Mechanism [\textbf{\#1}$\sim$\textbf{\#2}].} Coarse-to-fine mechanism is essential for scalable camera relocalization. Removing either stage inevitably degrades performance, highlighting contributions and complementary roles of these two stages.

\noindent\textbf{Scene Graph [\textbf{\#3}].} Scene graph analysis targets solving uncertain candidates in $U$. Without it, the system may be confused by similar or repeated objects.

\noindent\textbf{Language Modality [\textbf{\#4}].} Under visual occlusion and noise, the visual modality alone is biased, whereas language descriptions informed by an LLM agent’s common-sense reasoning can effectively fix this weakness.

\noindent\textbf{DIOU-based Retrieval [\textbf{\#5}].} We replace the DIOU-based retrieval with a naive visibility-based strategy. The results show that DIOU-based retrieval yields more reliable pose priors, thereby stabilizing pose estimation.

\noindent\textbf{Invalid Object Filtering [\textbf{\#6}].} Invalid objects disrupt landmark association and distort the scene graph, as ubiquitous elements (\eg, walls or floors) connect to most nodes. We find that retaining only informative objects is helpful.

\vspace{-0.5em}
\section{Conclusion}
\label{sec:conclusion}

We propose OpenReLoc, a comprehensive indoor camera relocalization system based on object-level representation, open-vocabulary understanding, and a dual-path 2D ICP loss. OpenReLoc is the first framework to handle scalable scenes in an object-level manner. Extensive experiments demonstrate that OpenReLoc achieves pioneering progress in object-level camera relocalization.

\textbf{Limitations.} 
Although scene graph analysis can help us address common indoor repetition cases, it struggles in extreme repetition, such as hundreds of identical chairs in an office environment. In addition, the efficiency of our system is currently affected by latency from online API calls of a closed-source LLM. We plan to improve and release a more efficient version with a locally deployed LLM in the future.

\vspace{-0.1em}


\bibliographystyle{IEEEtran}
\bibliography{main}

\begin{thebibliography}{10}
\providecommand{\url}[1]{#1}
\csname url@samestyle\endcsname
\providecommand{\newblock}{\relax}
\providecommand{\bibinfo}[2]{#2}
\providecommand{\BIBentrySTDinterwordspacing}{\spaceskip=0pt\relax}
\providecommand{\BIBentryALTinterwordstretchfactor}{4}
\providecommand{\BIBentryALTinterwordspacing}{\spaceskip=\fontdimen2\font plus
\BIBentryALTinterwordstretchfactor\fontdimen3\font minus \fontdimen4\font\relax}
\providecommand{\BIBforeignlanguage}[2]{{%
\expandafter\ifx\csname l@#1\endcsname\relax
\typeout{** WARNING: IEEEtran.bst: No hyphenation pattern has been}%
\typeout{** loaded for the language `#1'. Using the pattern for}%
\typeout{** the default language instead.}%
\else
\language=\csname l@#1\endcsname
\fi
#2}}
\providecommand{\BIBdecl}{\relax}
\BIBdecl

\bibitem{coordinet}
A.~Moreau, N.~Piasco, D.~Tsishkou, B.~Stanciulescu, and A.~de~La~Fortelle, ``Coordinet: uncertainty-aware pose regressor for reliable vehicle localization,'' in \emph{Proceedings of the IEEE/CVF Winter Conference on Applications of Computer Vision}, 2022, pp. 2229--2238.

\bibitem{pixloc}
P.-E. Sarlin, A.~Unagar, M.~Larsson, H.~Germain, C.~Toft, V.~Larsson, M.~Pollefeys, V.~Lepetit, L.~Hammarstrand, F.~Kahl \emph{et~al.}, ``Back to the feature: Learning robust camera localization from pixels to pose,'' in \emph{Proceedings of the IEEE/CVF conference on computer vision and pattern recognition}, 2021, pp. 3247--3257.

\bibitem{mur2017orb}
R.~Mur-Artal and J.~D. Tard{\'o}s, ``Orb-slam2: An open-source slam system for monocular, stereo, and rgb-d cameras,'' \emph{IEEE transactions on robotics}, vol.~33, no.~5, pp. 1255--1262, 2017.

\bibitem{ms-transformer}
Y.~Shavit, R.~Ferens, and Y.~Keller, ``Learning multi-scene absolute pose regression with transformers,'' in \emph{Proceedings of the IEEE/CVF International Conference on Computer Vision}, 2021, pp. 2733--2742.

\bibitem{zins2022oa}
M.~Zins, G.~Simon, and M.-O. Berger, ``Oa-slam: Leveraging objects for camera relocalization in visual slam,'' in \emph{2022 IEEE international symposium on mixed and augmented reality (ISMAR)}.\hskip 1em plus 0.5em minus 0.4em\relax IEEE, 2022, pp. 720--728.

\bibitem{goreloc}
Y.~Wang, C.~Jiang, and X.~Chen, ``Goreloc: Graph-based object-level relocalization for visual slam,'' \emph{IEEE Robotics and Automation Letters}, 2024.

\bibitem{peng2023openscene}
S.~Peng, K.~Genova, C.~Jiang, A.~Tagliasacchi, M.~Pollefeys, T.~Funkhouser \emph{et~al.}, ``Openscene: 3d scene understanding with open vocabularies,'' in \emph{Proceedings of the IEEE/CVF conference on computer vision and pattern recognition}, 2023, pp. 815--824.

\bibitem{clip}
A.~Radford, J.~W. Kim, C.~Hallacy, A.~Ramesh, G.~Goh, S.~Agarwal, G.~Sastry, A.~Askell, P.~Mishkin, J.~Clark \emph{et~al.}, ``Learning transferable visual models from natural language supervision,'' in \emph{International conference on machine learning}.\hskip 1em plus 0.5em minus 0.4em\relax PmLR, 2021, pp. 8748--8763.

\bibitem{scannet}
A.~Dai, A.~X. Chang, M.~Savva, M.~Halber, T.~Funkhouser, and M.~Nie{\ss}ner, ``Scannet: Richly-annotated 3d reconstructions of indoor scenes,'' in \emph{Proceedings of the IEEE conference on computer vision and pattern recognition}, 2017, pp. 5828--5839.

\bibitem{scannet++}
C.~Yeshwanth, Y.-C. Liu, M.~Nie{\ss}ner, and A.~Dai, ``Scannet++: A high-fidelity dataset of 3d indoor scenes,'' in \emph{Proceedings of the IEEE/CVF International Conference on Computer Vision}, 2023, pp. 12--22.

\bibitem{habitat19iccv}
M.~Savva, A.~Kadian, O.~Maksymets, Y.~Zhao, E.~Wijmans, B.~Jain, J.~Straub, J.~Liu, V.~Koltun, J.~Malik, D.~Parikh, and D.~Batra, ``Habitat: {A} {P}latform for {E}mbodied {AI} {R}esearch,'' in \emph{Proceedings of the IEEE/CVF International Conference on Computer Vision (ICCV)}, 2019.

\bibitem{jia2021scaling}
C.~Jia, Y.~Yang, Y.~Xia, Y.-T. Chen, Z.~Parekh, H.~Pham, Q.~Le, Y.-H. Sung, Z.~Li, and T.~Duerig, ``Scaling up visual and vision-language representation learning with noisy text supervision,'' in \emph{International conference on machine learning}.\hskip 1em plus 0.5em minus 0.4em\relax PMLR, 2021, pp. 4904--4916.

\bibitem{takmaz2023openmask3d}
A.~Takmaz, E.~Fedele, R.~W. Sumner, M.~Pollefeys, F.~Tombari, and F.~Engelmann, ``Openmask3d: open-vocabulary 3d instance segmentation,'' in \emph{Proceedings of the 37th International Conference on Neural Information Processing Systems}, 2023, pp. 68\,367--68\,390.

\bibitem{nguyen2024open3dis}
P.~Nguyen, T.~D. Ngo, E.~Kalogerakis, C.~Gan, A.~Tran, C.~Pham, and K.~Nguyen, ``Open3dis: Open-vocabulary 3d instance segmentation with 2d mask guidance,'' in \emph{Proceedings of the IEEE/CVF Conference on Computer Vision and Pattern Recognition}, 2024, pp. 4018--4028.

\bibitem{yan2024maskclustering}
M.~Yan, J.~Zhang, Y.~Zhu, and H.~Wang, ``Maskclustering: View consensus based mask graph clustering for open-vocabulary 3d instance segmentation,'' in \emph{Proceedings of the IEEE/CVF Conference on Computer Vision and Pattern Recognition}, 2024, pp. 28\,274--28\,284.

\bibitem{lu2023ovir}
S.~Lu, H.~Chang, E.~P. Jing, A.~Boularias, and K.~Bekris, ``Ovir-3d: Open-vocabulary 3d instance retrieval without training on 3d data,'' in \emph{Conference on Robot Learning}.\hskip 1em plus 0.5em minus 0.4em\relax PMLR, 2023, pp. 1610--1620.

\bibitem{yang2019cubeslam}
S.~Yang and S.~Scherer, ``Cubeslam: Monocular 3-d object slam,'' \emph{IEEE Transactions on Robotics}, vol.~35, no.~4, pp. 925--938, 2019.

\bibitem{clip-loc}
S.~Matsuzaki, T.~Sugino, K.~Tanaka, Z.~Sha, S.~Nakaoka, S.~Yoshizawa, and K.~Shintani, ``Clip-loc: Multi-modal landmark association for global localization in object-based maps,'' in \emph{2024 IEEE International Conference on Robotics and Automation (ICRA)}.\hskip 1em plus 0.5em minus 0.4em\relax IEEE, 2024, pp. 13\,673--13\,679.

\bibitem{CLIP-Clique}
S.~Matsuzaki, K.~Tanaka, and K.~Shintani, ``Clip-clique: Graph-based correspondence matching augmented by vision language models for object-based global localization,'' \emph{IEEE Robotics and Automation Letters}, 2024.

\bibitem{zeng20173dmatch}
A.~Zeng, S.~Song, M.~Nie{\ss}ner, M.~Fisher, J.~Xiao, and T.~Funkhouser, ``3dmatch: Learning local geometric descriptors from rgb-d reconstructions,'' in \emph{Proceedings of the IEEE conference on computer vision and pattern recognition}, 2017, pp. 1802--1811.

\bibitem{yolo}
J.~Redmon, S.~Divvala, R.~Girshick, and A.~Farhadi, ``You only look once: Unified, real-time object detection,'' in \emph{Proceedings of the IEEE conference on computer vision and pattern recognition}, 2016, pp. 779--788.

\bibitem{khanna2024habitat}
M.~Khanna, Y.~Mao, H.~Jiang, S.~Haresh, B.~Shacklett, D.~Batra, A.~Clegg, E.~Undersander, A.~X. Chang, and M.~Savva, ``Habitat synthetic scenes dataset (hssd-200): An analysis of 3d scene scale and realism tradeoffs for objectgoal navigation,'' in \emph{Proceedings of the IEEE/CVF Conference on Computer Vision and Pattern Recognition}, 2024, pp. 16\,384--16\,393.

\bibitem{tumrgbd}
J.~Sturm, N.~Engelhard, F.~Endres, W.~Burgard, and D.~Cremers, ``A benchmark for the evaluation of rgb-d slam systems,'' in \emph{Proc. of the International Conference on Intelligent Robot Systems (IROS)}, Oct. 2012.

\bibitem{zhu2022nice}
Z.~Zhu, S.~Peng, V.~Larsson, W.~Xu, H.~Bao, Z.~Cui, M.~R. Oswald, and M.~Pollefeys, ``Nice-slam: Neural implicit scalable encoding for slam,'' in \emph{Proceedings of the IEEE/CVF conference on computer vision and pattern recognition}, 2022, pp. 12\,786--12\,796.

\end{thebibliography}

\end{document}